\documentclass[conference]{IEEEtran}

\usepackage{cite}
\usepackage{amsmath,amssymb,amsfonts}
\usepackage{algorithmic}
\usepackage{graphicx}
\usepackage{textcomp}
\usepackage{booktabs}
\usepackage{multirow}
\usepackage[table]{xcolor}
\newcommand{\cmark}{\textcolor{green!60!black}{\ding{51}}}  
\newcommand{\xmark}{\textcolor{red!70!black}{\ding{55}}}    
\usepackage{hyperref}
\usepackage{amssymb}
\usepackage{newunicodechar}
\usepackage{booktabs}
\usepackage{pifont}
\newunicodechar{✓}{\checkmark}
\usepackage{xcolor}
\def\BibTeX{{\rm B\kern-.05em{\sc i\kern-.025em b}\kern-.08em
    T\kern-.1667em\lower.7ex\hbox{E}\kern-.125emX}}
\begin{document}

\title{FOCUS: Fused Observation of Channels for Unveiling Spectra}

\author{
\IEEEauthorblockN{
Xi Xiao\IEEEauthorrefmark{1},
Aristeidis Tsaris\IEEEauthorrefmark{2},
Anika Tabassum\IEEEauthorrefmark{2},
John Lagergren\IEEEauthorrefmark{2},
Larry M. York\IEEEauthorrefmark{2},
Tianyang Wang\IEEEauthorrefmark{1},
Xiao Wang\IEEEauthorrefmark{2}
}

\IEEEauthorblockA{\IEEEauthorrefmark{1}University of Alabama at Birmingham, Birmingham, AL, USA\\
xxiao@uab.edu, tw2@uab.edu}

\IEEEauthorblockA{\IEEEauthorrefmark{2}Oak Ridge National Laboratory, Oak Ridge, TN, USA\\
tsarisa@ornl.gov, tabassuma@ornl.gov, lagergrenjh@ornl.gov, yorklm@ornl.gov, wangx2@ornl.gov}

\IEEEauthorblockA{Corresponding author: Xiao Wang, wangx2@ornl.gov}
}

\maketitle

\begin{abstract}

Hyperspectral imaging (HSI) provides critical details for biology and agriculture, yet interpreting Vision Transformers (ViTs) on such high-dimensional data remains a challenge. Existing methods fail for two reasons: the computational cost of processing hundreds of bands is prohibitive, and models often suffer from ``attention collapse,'' where focus shifts to irrelevant tokens rather than meaningful spectral signals. We propose FOCUS, a framework that enables efficient spatial-spectral interpretability for frozen ViTs. Our method introduces two lightweight components: class-specific spectral prompts that anchor attention to relevant wavelength groups, and a learnable [SINK] token that actively filters out noisy attention. Unlike prior approaches, FOCUS generates 3D saliency maps in a single forward pass without requiring gradients or retraining the backbone. Experiments show that our method improves band-level IoU by 15\% and reduces attention collapse by 40\%. With less than 1\% parameter overhead, FOCUS makes interpretable HSI practical for real-world applications.

\end{abstract}


\section{Introduction}
\label{sec:intro}

Hyperspectral imaging (HSI) captures dense spectral signatures across hundreds of narrow, contiguous bands. This capability enables fine-grained analysis of physiological traits, such as stress, disease, and pigment content. Consequently, HSI has become a valuable tool in biology, agriculture, and environmental monitoring. Recently, Vision Transformers (ViTs) have achieved strong performance in hyperspectral recognition tasks by modeling long-range spatial and spectral dependencies.

However, this predictive power has not translated into interpretability. For real-world deployment, HSI models must explain not only ``where'' decisions are made (spatial saliency) but also ``which wavelengths'' are responsible. Current ViT interpretability tools fail to meet this requirement. Existing methods, such as Grad-CAM~\cite{gradcam}, LayerCAM~\cite{layercam}, and Prompt-CAM~\cite{promptcam}, are designed for 3-channel RGB inputs and do not scale to high-dimensional hyperspectral data. They typically ignore the spectral axis or produce coarse spatial maps without specific wavelength information, making them biologically uninformative. The main barrier is computational. Unlike RGB images, hyperspectral cubes often contain 200 to 300 channels, creating input token sequences that are orders of magnitude larger. Since ViT attention scales quadratically with the number of tokens, calculating full spectral interpretability is often intractable due to high memory and runtime costs~\cite{xiao2026not}. As a result, existing methods usually avoid the full spectrum, which leaves a fundamental gap in saliency modeling.

\begin{figure}[t]
  \centering
  \includegraphics[width=0.98\linewidth]{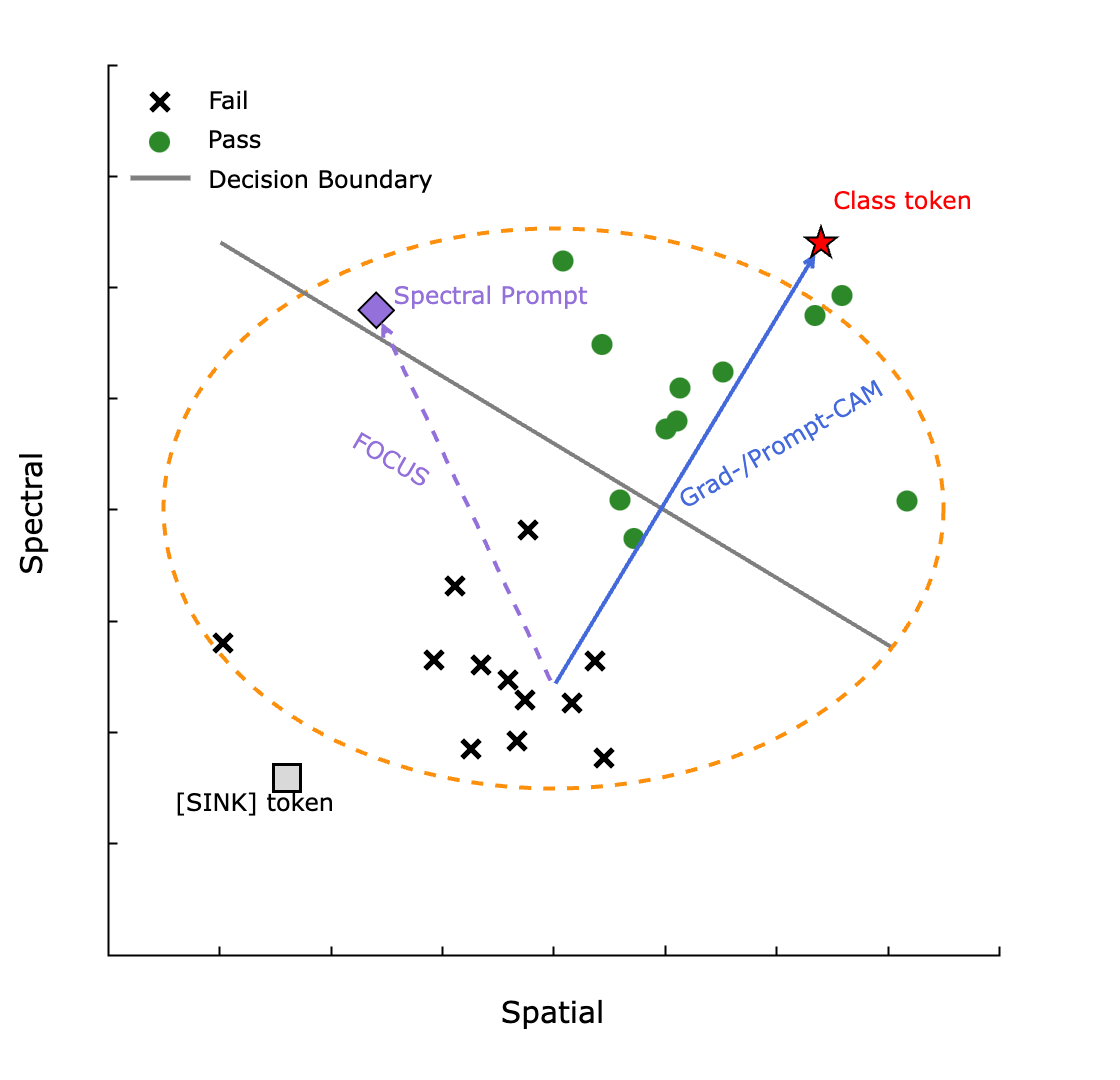}
  \caption{\textbf{Motivation of FOCUS.}  
  Black {$\times$} and green denote two classes of leaf cubes in a simplified spatial–spectral feature plane.}
  \label{fig:motivation}
\end{figure}

This challenge is further complicated by the \textit{attention sink} phenomenon~\cite{xiao2023efficient}. In this phenomenon, attention weights collapse onto dominant tokens, such as the \texttt{[CLS]} token. This issue becomes more severe in HSI as the spectral scale increases. With more spectral tokens, the attention becomes more diluted and unstable. When ViTs fail to capture meaningful wavelength patterns, they tend to revert to class tokens, causing spatial drift and spectral blindness (Fig.~\ref{fig:motivation}). For instance, in plant disease detection, key reflectance signals often reside in narrow bands like the red-edge (700–750\,nm). Existing methods fail to highlight these bands and produce diffuse saliency maps that do not align with expert knowledge, limiting their practical utility.

We address these challenges with \textbf{FOCUS}, the first framework to provide efficient and stable spatial–spectral interpretability for frozen ViTs on hyperspectral data. FOCUS introduces two lightweight innovations. First, we design \textbf{class-specific spectral prompts}. These prompts encode physiologically meaningful wavelength groups, such as VIS, red-edge, and NIR, to anchor attention in the spectral domain. Second, we introduce a learnable \textbf{\texttt{[SINK]} token} trained with an attraction loss. This token explicitly absorbs redundant or low-relevance attention, which turns the attention sink from a failure mode into a controlled filtering mechanism.

With these components, FOCUS enables reliable interpretation of ViT-based HSI models with minimal overhead. It produces a high-resolution 3D saliency cube across spatial and spectral dimensions, interpretable spectral attribution curves, and stable attention pathways. Importantly, it achieves this without requiring test-time gradients or backbone modifications. Extensive evaluations on two HSI benchmarks, including the new HyperLeaf2024 dataset, show that FOCUS improves band-level IoU by 15\%, reduces attention collapse by 40\%, and generates domain-aligned explanations. These gains are achieved with less than 1\% additional parameters.

To summarize, our contributions are:
\begin{itemize}
    \item We identify and resolve a fundamental scalability barrier for ViT-based interpretability in hyperspectral imaging. Existing methods cannot scale to full-spectrum inputs because attention computation grows exponentially with the number of spectral channels. Our framework offers the first practical solution for interpretable ViTs under high-dimensional spectral conditions.

    \item We introduce \textbf{FOCUS}, an interpretability framework that addresses both spectral relevance and attention stability. By combining class-specific spectral prompts and a learnable \texttt{[SINK]} token with an attraction loss, FOCUS enables precise wavelength attribution. It also effectively filters noisy or redundant attention, which substantially mitigates the attention collapse phenomenon.

    \item FOCUS is fully gradient-free, architecture-agnostic, and compatible with frozen ViTs. It generates high-resolution spatial–spectral explanations through a single forward pass and achieves state-of-the-art interpretability with less than 1\% parameter overhead. This makes it practical for real-world deployment in biology and agriculture.
\end{itemize}

\section{Method}
\label{sec:method}

\subsection{Overview}

Given a hyperspectral image cube $\mathbf{X} \in \mathbb{R}^{C \times H \times W}$ with $C$ spectral bands and spatial resolution $H \times W$, our goal is twofold: (i) accurately predict a target outcome, such as plant genotype, disease status, or physiological traits, and (ii) generate an interpretable \emph{joint spatial–spectral saliency map}, clearly indicating both \emph{where} and \emph{which wavelengths} influence model decisions. To achieve these objectives, we extend a frozen Vision Transformer (ViT)~\cite{vit} backbone by introducing three novel and intuitive modules:

\begin{enumerate}
    \item Class-specific spectral prompts, explicitly guiding wavelength-aware attention (Section~\ref{sec:prompt});
    \item A learnable \texttt{[SINK]} token with an associated attraction loss to convert attention collapse~\cite{xiao2023efficient} into a controllable noise-filtering mechanism (Section~\ref{sec:sink});
    \item An attention aggregation routine, forming a unified three-dimensional saliency cube from prompt-to-patch attention (Section~\ref{sec:cam}).
\end{enumerate}

\begin{figure*}[t]
  \centering
  \includegraphics[width=0.8\linewidth]{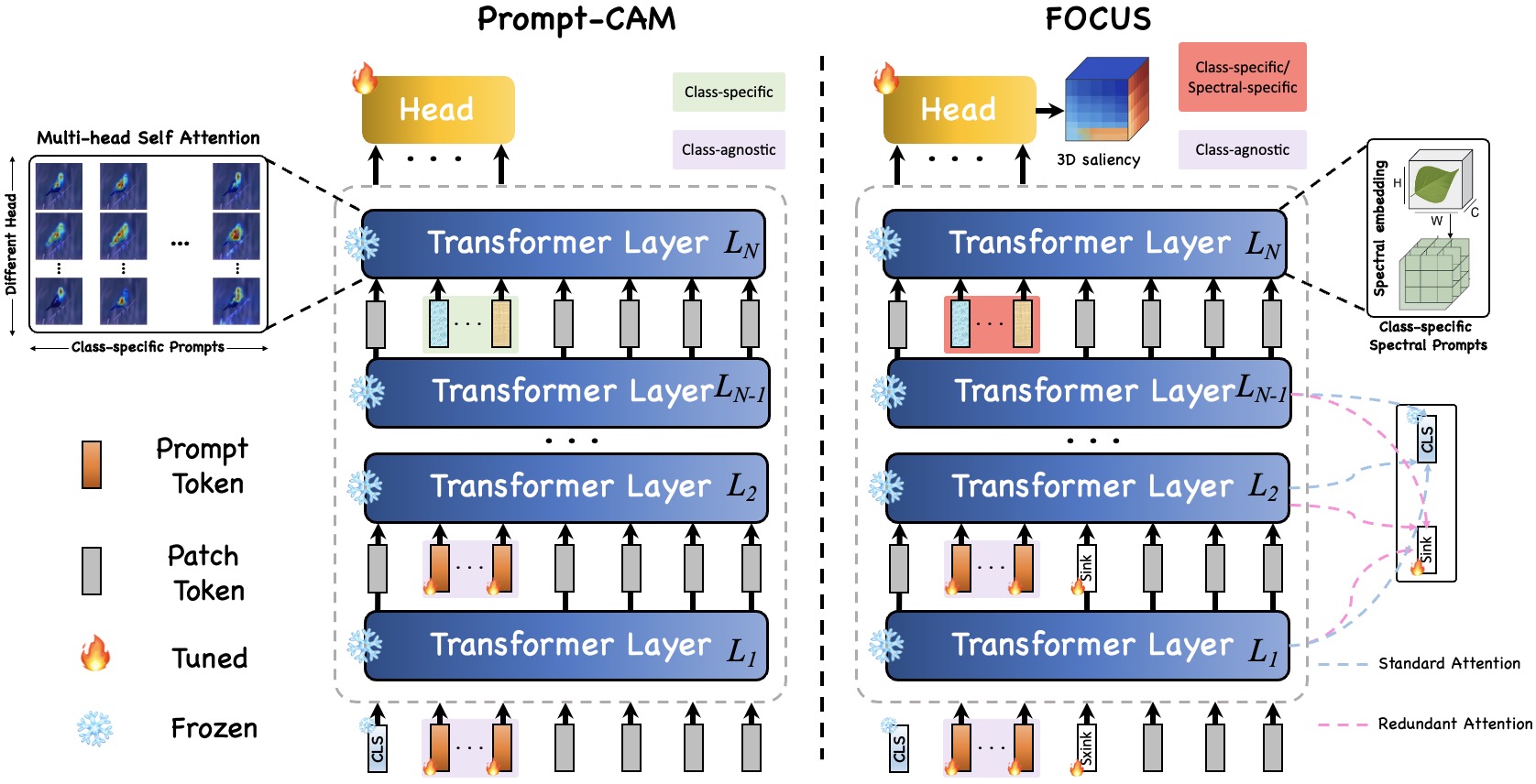}
  \caption{\textbf{Overview of \textsc{FOCUS}.} 
A hyperspectral image is embedded by a frozen ViT backbone. 
FOCUS augments it with (i) class-specific spectral prompts that anchor meaningful band groups (e.g., VIS, red-edge, NIR) and 
(ii) a dedicated \texttt{[SINK]} token to absorb redundant attention. 
Prompt-to-patch attention is aggregated into a 3D saliency cube $\mathbf{T} \in \mathbb{R}^{H \times W \times C}$, enabling joint spatial–spectral interpretation. 
We emphasize that the comparison with Prompt-CAM is only for contrast; the figure itself focuses on the FOCUS framework.}

  \label{fig:sprocam_overview}
\end{figure*}

Only the spectral prompts, \texttt{[SINK]} token, and a lightweight spectral adapter are trainable, collectively adding fewer than 1\% parameters to the ViT. Importantly, inference is fully gradient-free, as illustrated in Figure~\ref{fig:sprocam_overview}.

\subsection{Class-Specific Spectral Prompts}
\label{sec:prompt}

We partition the $C$ spectral bands into $G$ non-overlapping groups $\{\mathcal{B}_g\}_{g=1}^{G}$, guided by domain knowledge~\cite{hong2022spectralformer}, such as visible (VIS: 400–700 nm), red-edge (700–750 nm), NIR (750–1100 nm), and SWIR ($>$1100 nm). Let $C_g = |\mathcal{B}_g|$ denote the number of bands in group $g$.

For each class $c\in\{1,\dots,K\}$ and spectral group $g$, we introduce a learnable spectral prompt token $\mathbf{p}_{c,g}\in\mathbb{R}^{d}$. The initial input sequence is:
\begin{align}
\mathbf{Z}^{(0)}=\left[\underbrace{\mathbf{s}}_{\text{\scriptsize[SINK]}},\;
\underbrace{\mathbf{p}_{c,1},\dots,\mathbf{p}_{c,G}}_{\text{spectral prompts}},\;
\underbrace{\mathbf{x}_1,\dots,\mathbf{x}_N}_{\text{patch tokens}}\right],
\end{align}
where $N=H\cdot W$ is the number of spatial patches. Each band of $\mathbf{X}$ is first projected by a depth-wise $1{\times}1$ adapter $\boldsymbol{\Phi}\in\mathbb{R}^{C\times d}$, then patchified into $\{\mathbf{x}_i\}_{i=1}^N$. Thus $\mathbf{Z}^{(0)}$ explicitly concatenates [SINK], class-specific prompts, and embedded patch tokens.

Each spectral prompt $\mathbf{p}_{c,g}$ acts as a ``spectral anchor,'' interacting only with patches from its assigned band group $\mathcal{B}_g$. This design enforces group-aware attention routing and enhances spectral interpretability compared to generic prompts~\cite{promptcam,hyperprompt}.

\subsection{Sink-Aware Attention Routing}
\label{sec:sink}

At transformer layer $\ell$ and head $h$, attention is computed as:
\begin{align}
\mathbf{A}^{\ell,h}=\operatorname{softmax}\left(\frac{\mathbf{Q}^{\ell,h}\mathbf{K}^{\ell,h\top}}{\sqrt{d}}\right),
\end{align}
where $\mathbf{Q},\mathbf{K}\in\mathbb{R}^{(1+G+N)\times d}$, and the [SINK] token index is $k_{\mathrm{sink}}=1$.

We identify a subset of low-informativeness heads $\mathcal{H}_{\text{aux}}$ to serve as noise absorbers. Concretely, we compute attention entropy per head and select the bottom $30\%$ in each layer as $\mathcal{H}_{\text{aux}}$. We then apply an attraction loss:
\begin{align}
\mathcal{L}_{\text{sink}}=-\lambda\sum_{\ell}\frac{1}{|\mathcal{H}_{\text{aux}}|}\sum_{h\in\mathcal{H}_{\text{aux}}}\operatorname{mean}_i\left[\mathbf{A}^{\ell,h}_{i,k_\mathrm{sink}}\right],
\end{align}
where $\lambda$ is a small scalar (e.g., $10^{-3}$). 

We further define the \emph{[SINK] attention mass} at layer $\ell$, head $h$:
\begin{align}
\mathrm{mass}^{\ell,h} = \tfrac{1}{N}\sum_{i=1}^N \mathbf{A}^{\ell,h}_{i,k_{\mathrm{sink}}},
\end{align}
with \emph{Sink-Rate} as the mean mass across layers/heads, and \emph{Sink-Consistency} as the average Spearman correlation of saliency distributions across layers. These metrics quantitatively measure collapse mitigation.

\subsection{3-D Prompt-CAM Generation}
\label{sec:cam}

After the final layer $L$, prompt-to-patch attention from discriminative heads $\mathcal{H}_\star=\{1,\dots,H\}\setminus\mathcal{H}_{\text{aux}}$ is aggregated:
\begin{align}
T_{x,y,\lambda}=\frac{1}{|\mathcal{H}_\star|}\sum_{h\in\mathcal{H}_\star}\mathbf{A}^{L,h}_{q=\mathbf{p}_{c,g(\lambda)},\,k=\mathbf{x}_{x,y,\lambda}},
\end{align}
where $g(\lambda)$ is the spectral group index of band $\lambda$. This yields a spatial–spectral cube $\mathbf{T}\in\mathbb{R}^{H\times W\times C}$.

We derive a spatial heatmap and spectral curve as:
\begin{align}
M_{x,y}=\sum_{\lambda}T_{x,y,\lambda}, \quad B_{\lambda}=\max_{x,y}T_{x,y,\lambda}.
\end{align}
Both are obtained in a single forward pass, without gradients.

\subsection{Training Objective and Complexity}
\label{sec:loss}

The total objective combines task-specific loss $\mathcal{L}_{\text{task}}$ (cross-entropy or MSE) and $\mathcal{L}_{\text{sink}}$:
\begin{align}
\mathcal{L}=\mathcal{L}_{\text{task}}+\mathcal{L}_{\text{sink}}.
\end{align}

During training, only (i) spectral prompts ($K\times G\times d$), (ii) the [SINK] token ($d$), and (iii) the $1\times1$ spectral adapter ($C\times d$) are updated, adding fewer than 1\% parameters (0.8M vs 86M for ViT-B/16) and $\sim$2.1\% FLOPs.

\section{Experiments}
\label{sec:experiments}

\subsection{Experimental Setup}
\label{sec:exp_setup}

\paragraph{Datasets}We evaluate FOCUS on two representative hyperspectral leaf datasets to comprehensively evaluate its performance in both classification and regression settings. The first, \textbf{HyperLeaf 2024}~\cite{hyperleaf2024}, comprises 2,410 VNIR--SWIR hyperspectral image cubes, each with a spatial resolution of $256{\times}256$. Each sample is annotated with both genotype labels (4 classes) and continuous physiological indicators such as chlorophyll \textsubscript{ab} and water content, enabling the simultaneous study of discrete classification and continuous trait regression. The second dataset, \textbf{Tomato BLS}~\cite{tomatobls}, contains 3,120 hyperspectral images of tomato leaves, labeled as either healthy or infected with bacterial leaf spot (BLS). This binary disease classification task is particularly suitable for evaluating the precision of saliency methods in localizing stress-affected regions and detecting spectrally discriminative cues under real-world noise and biological variation.

\paragraph{Evaluation Metrics}We adopt a suite of established metrics that collectively capture classification accuracy, spectral attribution fidelity, and spatial alignment with expert annotations. \textbf{Classification Accuracy (Acc)} measures standard top-1 prediction correctness. \textbf{Band-level IoU (BIO@5)}, introduced by SANet~\cite{sanet}, quantifies spectral interpretability by computing Intersection-over-Union over the top-5 most influential wavelengths. \textbf{Spatial IoU}, following the DiffCAM protocol~\cite{diffcam}, evaluates overlap between predicted saliency regions and ground-truth lesion masks. Lastly, \textbf{AUPRC (Area Under Precision--Recall Curve)} is used to evaluate fine-grained spatial localization quality, especially in class-imbalanced disease identification scenarios. This evaluation protocol aligns with prevailing practices in the hyperspectral interpretability literature~\cite{sanet,diffcam,promptcam}. All results are averaged over three random dataset splits to ensure robustness.

\begin{figure}[t]
  \centering
  \includegraphics[width=0.95\linewidth]{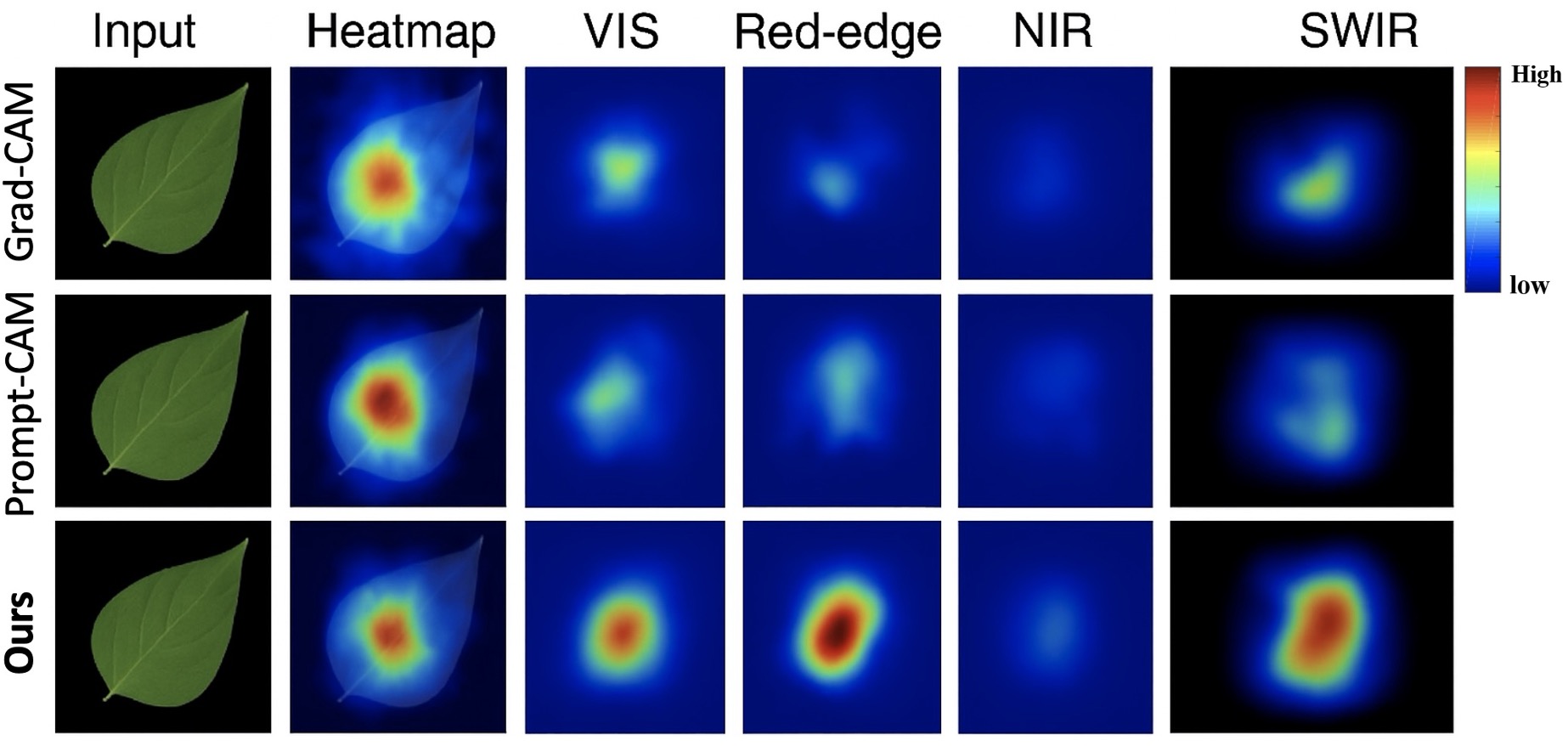}
  \caption{\textbf{Band-wise interpretability across spectral groups.}
  Compared to baseline methods (Grad-CAM and Prompt-CAM), \textbf{FOCUS} produces significantly clearer and biologically grounded saliency patterns, especially in the red-edge and shortwave infrared (SWIR) regions (columns 4–5).
  The red-edge band (700–750\,nm) is known to reflect changes in chlorophyll concentration and leaf pigment activity, critical for identifying early-stage stress and disease.
  The SWIR band (around 2,100\,nm) is highly sensitive to water content and cellular structure, offering key cues for detecting drought or tissue damage.
  Our method highlights these regions with greater clarity and spatial precision, aligning closely with expert physiological annotations and enhancing model transparency in plant health assessment.}
  \label{fig:qual_saliency3}
\end{figure}

\subsection{Main Results}
\label{sec:exp_main}

We compare FOCUS against competitive baselines on two challenging hyperspectral benchmarks: \textbf{HyperLeaf 2024} and \textbf{Tomato BLS}. The quantitative results are summarized in Table~\ref{tab:multi_dataset_results}. For classification, we evaluate performance using Accuracy (Acc), band-level interpretability (BIO@5), spatial IoU, and AUPRC. For regression tasks on HyperLeaf2024, we report Mean Squared Error (MSE) and $R^2$. To ensure fair comparison, all methods utilize the same frozen ViT-B/16 (DINOv2) backbone. As shown in the table, FOCUS consistently outperforms baselines across all metrics. It demonstrates not only strong predictive accuracy but also superior interpretability in both spatial and spectral dimensions. On HyperLeaf 2024, our method achieves substantial gains. Specifically, it improves BIO@5 and IoU by over 15\% compared to the best transformer-based alternatives. Similarly, on Tomato BLS, FOCUS yields sharper lesion localization and more accurate band attribution. This highlights its robustness against domain shifts caused by disease. The advantages extend to regression tasks as well. FOCUS reduces MSE by approximately 10--15\% and increases $R^2$ by 5--8 points across all targets. Collectively, these results validate FOCUS as a robust and unified framework for accurate HSI modeling.

\begin{table*}[t]
\centering
\renewcommand{\arraystretch}{1.05}
\setlength{\tabcolsep}{3pt}
\caption{
\textbf{Main results across hyperspectral benchmarks.}
For classification (HyperLeaf2024, Tomato BLS), we report accuracy (Acc), BIO@5, IoU, and AUPRC. 
For regression (HyperLeaf2024: GrainWeight, $G_{sw}$, $\Phi_{\mathrm{PSII}}$), we report mean squared error (MSE, $\downarrow$) and $R^2$ ($\uparrow$). 
Best results are in \textbf{bold}.
}
\vspace{2pt}
\scriptsize
\begin{tabular}{l|cccc|cccc|cc|cc|cc}
\toprule
\multirow{2}{*}{\textbf{Method}} 
& \multicolumn{4}{c|}{\textbf{HyperLeaf2024 (Cls)}} 
& \multicolumn{4}{c|}{\textbf{Tomato BLS (Cls)}} 
& \multicolumn{2}{c|}{\textbf{GrainWeight}} 
& \multicolumn{2}{c|}{\textbf{$G_{sw}$}} 
& \multicolumn{2}{c}{\textbf{$\Phi_{\mathrm{PSII}}$}} \\
\cmidrule(lr){2-5} \cmidrule(lr){6-9} \cmidrule(lr){10-11} \cmidrule(lr){12-13} \cmidrule(lr){14-15}
& Acc & BIO@5 & IoU & AUPRC 
& Acc & BIO@5 & IoU & AUPRC
& MSE & $R^2$ 
& MSE & $R^2$ 
& MSE & $R^2$ \\
\midrule
Grad-CAM~\cite{gradcam}{\scriptsize [ICCV'17]}    
& 82.1 & 0.33 & 0.41 & 0.53 
& 79.3 & 0.28 & 0.39 & 0.51 
& 0.142 & 0.62 & 0.158 & 0.59 & 0.167 & 0.57 \\
SANet~\cite{sanet}{\scriptsize [TIP'20]}          
& 83.0 & 0.36 & 0.43 & 0.56 
& 80.5 & 0.29 & 0.41 & 0.52 
& 0.138 & 0.64 & 0.152 & 0.61 & 0.161 & 0.60 \\
Prompt-CAM~\cite{promptcam}{\scriptsize [CVPR'25]}  
& 84.5 & 0.38 & 0.46 & 0.59 
& 81.2 & 0.34 & 0.43 & 0.58 
& 0.134 & 0.66 & 0.148 & 0.63 & 0.156 & 0.62 \\
DiffCAM~\cite{diffcam}{\scriptsize [CVPR'25]}      
& 85.3 & 0.41 & 0.49 & 0.62 
& 82.4 & 0.36 & 0.47 & 0.61 
& 0.128 & 0.69 & 0.145 & 0.65 & 0.150 & 0.64 \\
\rowcolor{blue!8}
\textbf{FOCUS (Ours)} 
& \textbf{87.0} & \textbf{0.56} & \textbf{0.58} & \textbf{0.72} 
& \textbf{84.6} & \textbf{0.47} & \textbf{0.54} & \textbf{0.70} 
& \textbf{0.115} & \textbf{0.75} & \textbf{0.132} & \textbf{0.72} & \textbf{0.141} & \textbf{0.70} \\
\bottomrule
\end{tabular}
\label{tab:multi_dataset_results}
\end{table*}

\begin{table}[t]
\centering
\scriptsize 
\setlength{\tabcolsep}{3pt} 
\renewcommand{\arraystretch}{1.0} 
\caption{\textbf{Efficiency and inference overhead.} 
FOCUS adds minimal overhead and requires no gradients.}
\vspace{2pt}
\begin{tabular}{lcccc}
\toprule
\textbf{Method} & Params (\%) & FLOPs (\%) & Grad & Latency (ms) \\
\midrule
Grad-CAM~\cite{gradcam}{\scriptsize [ICCV'17]}   & 0.0  & 12.5 & \cmark & 11.2 \\
SANet~\cite{sanet}{\scriptsize [TIP'20]}         & 2.1  & 8.2  & \cmark & 10.5 \\
Prompt-CAM~\cite{promptcam}{\scriptsize [CVPR'25]} & 1.3  & 5.6  & \xmark & 8.6 \\
DiffCAM~\cite{diffcam}{\scriptsize [CVPR'25]}    & 0.5  & 4.8  & \xmark & 7.9 \\
\rowcolor{blue!8}
\textbf{FOCUS (Ours)} & \textbf{0.9} & \textbf{2.1} & \textbf{\xmark} & \textbf{6.8} \\
\bottomrule
\end{tabular}
\label{tab:efficiency}
\end{table}

\paragraph{Efficiency and Inference Overhead.}
As shown in Table~\ref{tab:efficiency}, FOCUS incurs less than 1\% parameter overhead and achieves the lowest FLOPs and runtime latency among all baselines. Importantly, it operates in a fully gradient-free manner at inference, which sharply reduces deployment cost compared to gradient-based methods (e.g., Grad-CAM~\cite{gradcam}, SANet~\cite{sanet}). These findings are consistent with recent work~\cite{promptcam,hyperprompt}, highlighting that prompt-based and gradient-free explainers enable practical, real-time visual interpretation for large transformer models.


\subsection{Ablation Study}
\label{sec:exp_ablation}

\paragraph{Spectral Prompt Granularity.}
Table~\ref{tab:prompt_granularity} shows that increasing the number of spectral groups ($G$) improves both accuracy and interpretability, but only up to a moderate value ($G=10$), beyond which gains saturate or slightly decline. This echoes prior findings~\cite{hyperprompt,sanet} that over-partitioning can dilute spectral focus and increase model overhead.

\paragraph{Sink Control Mechanism.}
Table~\ref{tab:sink_ablation} demonstrates that introducing a dedicated \texttt{[SINK]} token and the associated attraction loss yields clear improvements in both interpretability and explanation stability, reducing spurious attention accumulation. This is consistent with recent work on attention regularization and prompt-based filtering in transformers~\cite{xiao2023efficient,darcet2024vision,hyperprompt}.

\subsection{Qualitative Analysis}
\label{sec:exp_qual}

\begin{figure}[t]
  \centering
  \includegraphics[width=0.79\linewidth]{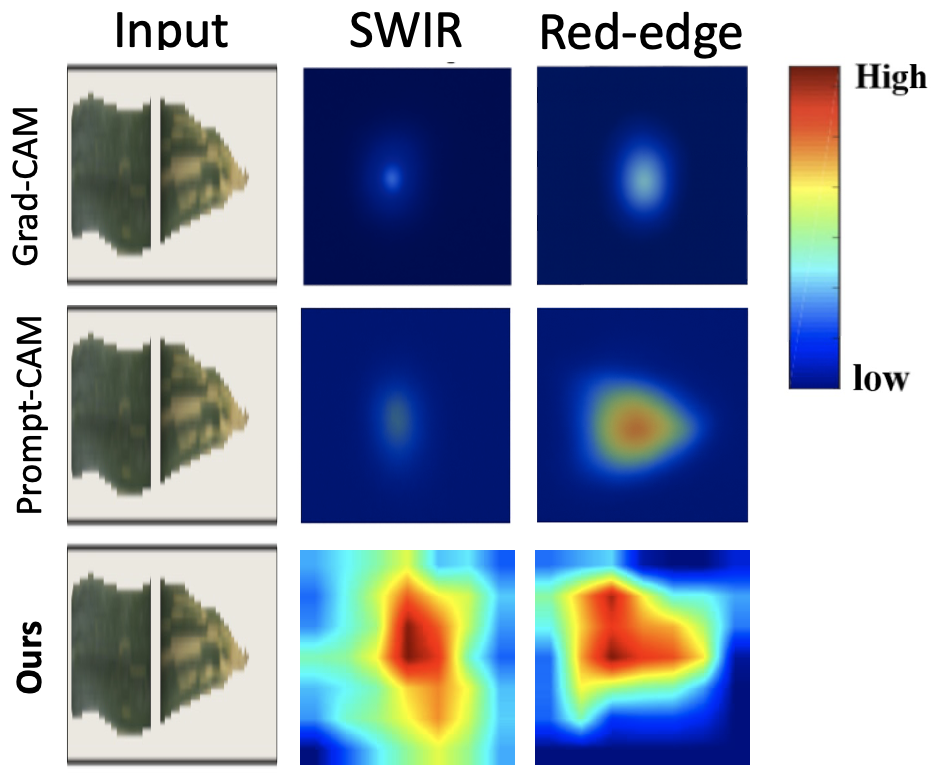}
  \caption{\textbf{Saliency maps across spatial and spectral dimensions.}
  Compared to Grad-CAM and Prompt-CAM, FOCUS exhibits both sharper spatial saliency and cleaner spectral separation, especially at red-edge (750\,nm) and SWIR (2,100\,nm) bands.}
  \label{fig:qual_saliency1}
\end{figure}

\begin{figure}[t]
  \centering
  \includegraphics[width=0.66\linewidth]{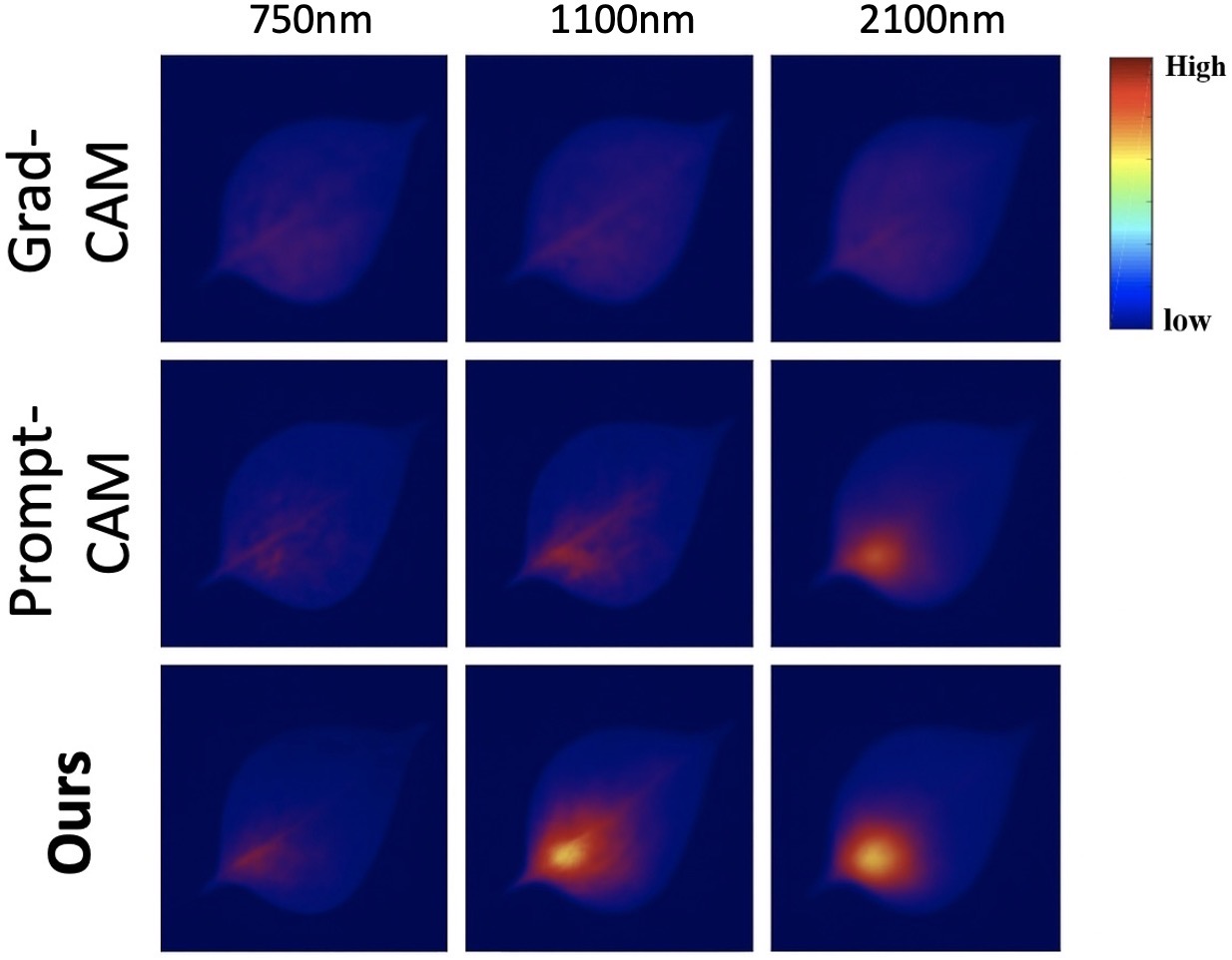}
  \caption{\textbf{Band-wise saliency comparison at key wavelengths.}
  Visualization of saliency maps at 750\,nm (red-edge), 1100\,nm (NIR), and 2100\,nm (SWIR).  
  Compared to Grad-CAM and Prompt-CAM, our method produces clearer and more concentrated saliency, particularly at red-edge and SWIR bands, which are known to be physiologically relevant for leaf health. The colorbar reflects normalized attention intensities.}
  \label{fig:band_saliency}
\end{figure}

We use visual comparisons to evaluate the interpretability and spectral coherence of FOCUS. Overall, FOCUS produces cleaner and more localized saliency patterns than existing methods. As shown in Figure~\ref{fig:qual_saliency3}, Grad-CAM tends to generate diffuse saliency maps that offer limited value under challenging spectral conditions. Prompt-CAM improves contrast but still struggles to achieve fine-grained localization in biologically informative regions. In contrast, FOCUS generates sharp and consistent attention maps. These maps align closely with known physiological markers. Specifically, the model accurately highlights stress-induced lesions in the spatial domain. It also identifies distinct spectral responses in the red-edge (750\,nm) and SWIR ($\sim$2,100\,nm) bands. These wavelengths are linked to chlorophyll degradation and protein absorption in plant tissues, which supports the biological validity of our approach. Figure~\ref{fig:qual_saliency1} demonstrates the robustness of FOCUS across replicates. The attention consistently targets biochemically informative areas. Conversely, Prompt-CAM exhibits unstable activation, and Grad-CAM fails to distinguish the object from background noise. This indicates that our spectral prompts and sink routing mechanism ensure more stable attention allocation. Finally, Figure~\ref{fig:band_saliency} displays band-wise heatmaps for different spectral groups. FOCUS identifies not only where important features lie spatially but also which specific bands are significant. For example, it reveals sharp peaks in the red-edge and protein-sensitive SWIR bands. In comparison, baseline methods often collapse saliency across bands or highlight spectrally irrelevant regions, which provides limited insight for domain experts. These qualitative findings reinforce our quantitative results and validate FOCUS as a reliable tool for interpretable hyperspectral vision.

\begin{table}[t]
\scriptsize
\centering
\setlength{\tabcolsep}{3pt}
\renewcommand{\arraystretch}{0.95}
\caption{\textbf{Sink token and loss.} Accuracy and explanation stability improve with explicit sink routing.}
\vspace{2pt}
\begin{tabular}{l|cccc}
\toprule
\textbf{Config} & Acc $\uparrow$ & BIO@5 $\uparrow$ & Sink Rate $\downarrow$ & Consist. $\uparrow$ \\
\midrule
w/o Sink        & 93.5 & 0.51 & 0.38 & 0.64 \\
+ Sink, no loss & 93.4 & 0.53 & 0.31 & 0.69 \\
\rowcolor{blue!8}
+ Sink \& Loss  & \textbf{93.7} & \textbf{0.56} & \textbf{0.22} & \textbf{0.75} \\
\bottomrule
\end{tabular}
\label{tab:sink_ablation}
\end{table}

\begin{table}[t]
\scriptsize 
\centering
\setlength{\tabcolsep}{3pt} 
\renewcommand{\arraystretch}{0.95} 
\caption{\textbf{Spectral prompt granularity.} $G{=}10$ achieves best interpretability without excess overhead.}
\vspace{2pt}
\begin{tabular}{c|cccc}
\toprule
\textbf{$G$} & Acc $\uparrow$ & BIO@5 $\uparrow$ & $\rho$ $\uparrow$ & Params (\%) \\
\midrule
2   & 91.1 & 0.48 & 0.63 & 0.4 \\
5   & 92.6 & 0.52 & 0.68 & 0.6 \\
10  & \textbf{93.7} & \textbf{0.56} & \textbf{0.72} & 0.9 \\
20  & 92.9 & 0.54 & 0.70 & 1.6 \\
\bottomrule
\end{tabular}
\label{tab:prompt_granularity}
\vspace{-4pt}
\end{table}

\subsection{Robustness Under Spectral Noise}
\label{sec:exp_noise}

Hyperspectral imaging is inherently susceptible to noise at high wavelengths (especially SWIR $>$2200\,nm) due to sensor artifacts, water absorption, or atmospheric effects. This not only reduces model reliability, but can also destabilize saliency explanations as transformers overfit to spurious spectral patterns. Empirically, we observe that vanilla ViT-based saliency methods (e.g., Grad-CAM~\cite{gradcam}, Prompt-CAM~\cite{promptcam}) often produce volatile or physiologically irrelevant attributions in these noisy bands—a phenomenon also documented in~\cite{zhou2021deepvit,chefer2022robustvit}. In contrast, our sink mechanism acts as a noise valve: unstable heads route high-entropy responses to \texttt{[SINK]}, selectively suppressing noise-induced activations. This yields more robust and biologically meaningful explanations, without additional model tuning. Our results resonate with the “attention debiasing” literature, which finds that targeted attention steering—whether by loss design~\cite{chefer2022robustvit} or prompt gating~\cite{baraldi2025mapet}—improves model resilience to common corruptions. Uniquely, FOCUS achieves these gains with less than 1\% additional parameters and without retraining the ViT backbone.

\subsection{Computational Efficiency Comparison}
\label{sec:exp_efficiency}

\begin{figure}[t]
  \centering
  \includegraphics[width=0.9\linewidth]{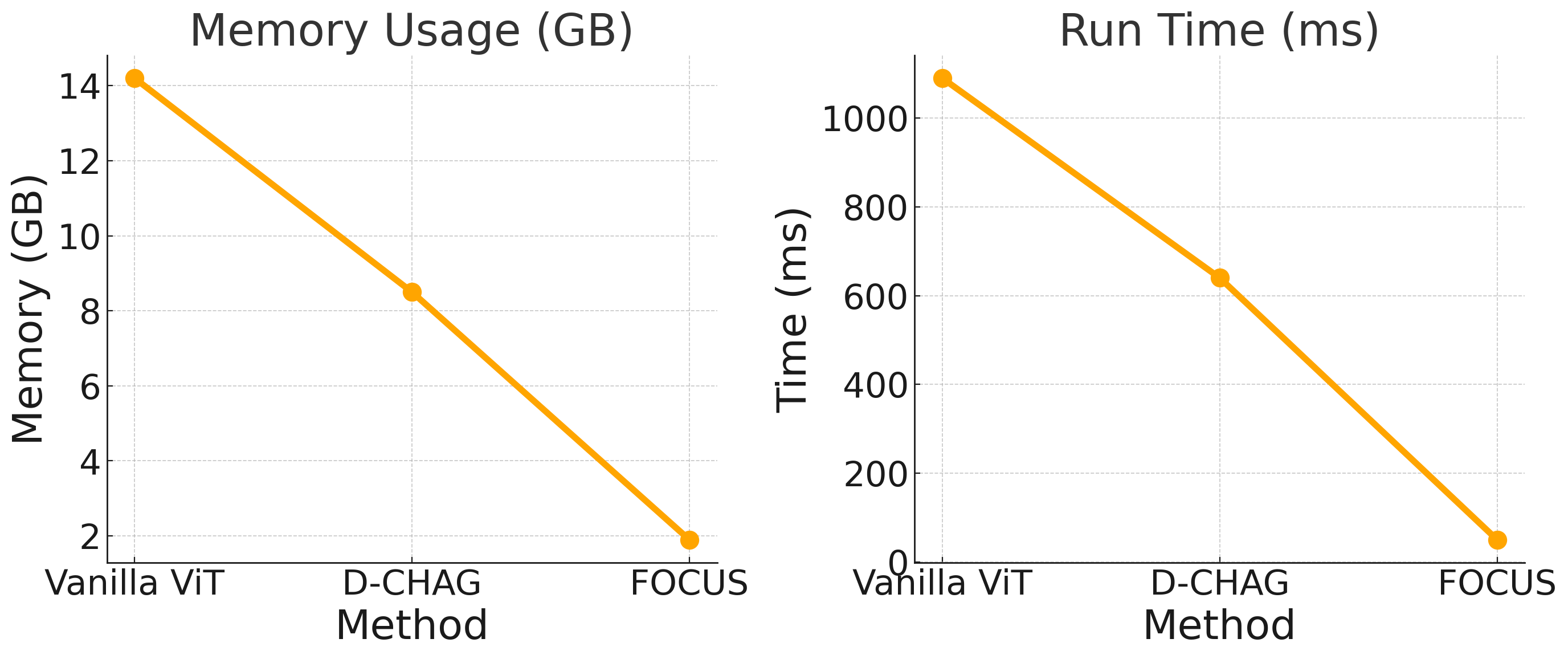}
  \caption{\textbf{Inference efficiency comparison.}
  FOCUS achieves over 95\% runtime and 85\% memory reduction compared to vanilla ViT and D-CHAG~\cite{tsaris2025distributed}, while providing complete spatial spectral interpretability.}
  \label{fig:runtime_memory_compare}
\end{figure}

Interpretability in hyperspectral vision must be efficient to support deployment in real-world agricultural and biological scenarios with limited resources. However, standard ViTs scale poorly for hyperspectral cubes: a single image may contain 200+ spectral bands, leading to quadratic growth in memory and attention cost. Direct inference with a vanilla ViT requires over 14,GB of GPU memory. Recent approaches like D-CHAG~\cite{tsaris2025distributed} use spectral aggregation to reduce computation, yet still consume 8.4,GB and require model modifications and retraining. In contrast, \textbf{FOCUS} introduces no backbone changes or gradients. Through lightweight spectral prompts and a learnable \texttt{[SINK]} token, it requires only 1.9,GB of memory and operates in a single forward pass. As shown in Figure~\ref{fig:runtime_memory_compare}, FOCUS reduces memory cost by over 85\% while delivering fine-grained spatial spectral explanations. This efficiency makes our method the first practically deployable interpretability solution for ViTs in high-dimensional hyperspectral vision.

\section{Conclusion}
\label{sec:conclusion}

We introduced \textbf{FOCUS}, a lightweight and gradient-free interpretability framework tailored for hyperspectral vision transformers. By combining class-specific spectral prompts with a learnable \texttt{[SINK]} token and a simple attraction loss, FOCUS produces stable and fine-grained spatial–spectral saliency maps. Despite adding less than 1\% additional parameters, it achieves state-of-the-art interpretability performance—improving spectral IoU by 15\% and enhancing attribution consistency across layers. Our work addresses a fundamental scalability barrier in hyperspectral interpretability, enabling trustworthy and deployable explanations for high-dimensional spectral data. We believe FOCUS lays a practical foundation for interpretable AI in spectral-intensive domains. In future work, we aim to explore adaptive prompt sharing and hierarchical strategies to further scale FOCUS to long-sequence spectral tasks.In the future, we can extend our work into reinforcement learning or multi-agent systems to further enhance the interpretability of hyperspectral images~\cite{duanadaptive,11059994}.

\section*{Acknowledgments}
This manuscript was co-authored by Oak Ridge National Laboratory (ORNL), operated by UT-Battelle, LLC under Contract No. DE-AC05-00OR22725 with the U.S. Department of Energy.  
Any subjective views or opinions expressed in this paper do not necessarily represent those of the U.S. Department of Energy or the United States Government.

\bibliographystyle{IEEEbib}
\bibliography{icme2026references}

\end{document}